\documentclass{article} 
\usepackage[accepted]{icml2018}
\usepackage{url}
\usepackage{times}  
\usepackage{helvet} 
\usepackage{amsmath}
\usepackage{color,framed}
\usepackage{courier}  
\usepackage{url}  
\usepackage{graphicx}  
\usepackage{booktabs}
\usepackage{wrapfig} 
\usepackage{multirow}

\usepackage{titlesec}
\titlespacing{\paragraph}{%
  0pt}{
  0em}{
  1em}

\begin{document}

\twocolumn[
\icmltitle{Focused Hierarchical RNNs for Conditional Sequence Processing}

\icmlsetsymbol{equal}{*}

\begin{icmlauthorlist}
\icmlauthor{Nan Rosemary Ke}{mila,poly,msr}
\icmlauthor{Konrad \.Zo\l{}na}{uj,mila}
\icmlauthor{Alessandro Sordoni}{msr}
\icmlauthor{Zhouhan Lin}{mila,msr,adept}
\icmlauthor{Adam Trischler}{msr}
\icmlauthor{Yoshua Bengio}{mila,udem,cifar}
\icmlauthor{Joelle Pineau} {mcgill,facebook,cifar}
\icmlauthor{Laurent Charlin}{mila,hec}
\icmlauthor{Chris Pal}{mila,poly}

\end{icmlauthorlist}

\icmlaffiliation{mila}{Montreal Institute for Learning Algorithms, Montreal, Canada}
\icmlaffiliation{poly}{Polytechnique Montreal}
\icmlaffiliation{msr}{Microsoft Research, Montreal}
\icmlaffiliation{uj}{Jagiellonian University, Cracow, Poland}
\icmlaffiliation{adept}{AdeptMind Scholar}
\icmlaffiliation{cifar}{Senior Cifar Member}
\icmlaffiliation{udem}{University of Montreal}
\icmlaffiliation{facebook}{Facebook AI Research, Montreal}
\icmlaffiliation{mcgill}{McGill University}
\icmlaffiliation{hec}{HEC Montreal}
\icmlcorrespondingauthor{Nan Ke}{nan.ke@polymtl.ca}

\icmlkeywords{Deep Learning, Recurrent Neural Networks, Attention Mechanism, Natural Language Processing}

\vskip 0.3in
]

\printAffiliationsAndNotice{}

\begin{abstract}
Recurrent Neural Networks (RNNs) with attention mechanisms have obtained state-of-the-art results for many sequence processing tasks. Most of these models use a simple form of encoder with attention that looks over the entire sequence and assigns a weight to each token independently.
We present a mechanism for focusing RNN encoders for sequence modelling tasks which allows them to attend to key parts of the input as needed. We formulate this using a multi-layer conditional 
sequence encoder that reads in one token at a time and makes a discrete decision on whether the token is relevant to the context or question being asked. The discrete gating mechanism takes in the context embedding and the current hidden state as inputs and controls information flow into the layer above. We train it using policy gradient methods. We evaluate this method on several types of tasks with different attributes. First, we evaluate the method on synthetic tasks which allow us to evaluate the model for its generalization ability and probe the behavior of the gates in more controlled settings.
We then evaluate this approach on large scale Question  Answering tasks including the challenging MS MARCO and SearchQA tasks. Our models shows consistent improvements for both tasks over prior work and our baselines. It has also shown to generalize significantly better on synthetic tasks as compared to the baselines.
\end{abstract}

\section{Introduction}
Recurrent Neural Networks (RNNs) with attention are wildly used for many sequence modeling tasks, such as: image captioning \cite{yao2016boosting,lu2017knowing}, speech recognition \cite{chan2016listen, bahdanau2016end}, text summarization \cite{nallapatiZSGX16} and Question and Answering (QA) \cite{KadlecSBK16}. The attention mechanism allows the model to look over the entire sequence and pick up the most relevant information.
This not only allows the model to learn a dynamic summarization of the input sequence, it allows gradients to be passed directly to the earlier time-steps in the input sequence, which also helps with the vanishing and exploding gradient problem \cite{Hochreiter91,BengioLLD94,Hochreiter98}.

Most of these models use a simple form of encoder with attention that is identical to the first one proposed \cite{bahdanau2014}, where the attention looks over the entire encoded sequence and assigns a soft weight to each token. However, for more complex tasks we conjecture that more structured encoding mechanisms may help the attention to more effectively identify and selectively process relevant information within the input.

Imagine reading a Wikipedia article and trying to identify information that is relevant to answering a question before one knows what the question is. Now, compare this to the situation where the context or question is given before reading the article. It would be much easier to read over the article, identify relevant information, group items and selectively process relevant information based on its relevance to a given context or question.

Keeping this intuition in mind, we have developed a focused RNN encoder that is modeled by a multi-layer RNN that groups input sub-sequences based on gates that are controlled or conditioned on a question or input context. We dub the core part of the minimal form of this model a \emph{focused hierarchical encoder} module. Our approach represents a general framework that applies to many sequence modeling tasks where the context or question can be beneficial to focus (attend) over the input sequence. 
Our focused encoder module examined here is based on a two layer LSTM where the upper layer is updated when a group of relevant tokens has been read. The boundaries of the group are computed using a discrete gating mechanism that takes as input the lower and upper level units as well as the context or question, and it is trained using policy gradient methods.

We evaluate our model on several tasks of different levels of complexity. We began with toy tasks with limited vocabulary size, where we analyze the performance, the generalization ability as well as the gating and the attention mechanisms. We then move on to challenging large scale QA tasks such as MS MARCO \citep{nguyen2016ms} and SearchQA \citep{dunn2017searchqa}. Our model outperforms the baseline for both tasks. For the SearchQA task, it significantly outperforms recently proposed methods \cite{buck2018ask}.

The key contributions of our work are the following:
\begin{itemize}
    \item We explore the use of a conditional discrete stochastic  boundary gating mechanism that helps the encoder to focus on parts relevant to the context, we use a soft attention to look over the relevant states.
    \item We use a reinforcement learning approach to learn the boundary gating mechanism.
    \item The elements above form the building blocks of our proposed \emph{focused hierarchical encoder} module, we examine its properties with synthetic data experiments and we show the benefits of using it for QA tasks.
\end{itemize}

Our model takes an input sequence, a question or context sequence and generates an answer. It can be applied to any sequence tasks where the context or question is beneficial to modulating the processing of an input sequence.

\section{Focused Hierarchical RNN}
\subsection{Architecture}\label{sec:passage_encoder}
\begin{figure}[!b]
    \centering
    \includegraphics[width=0.9\linewidth]{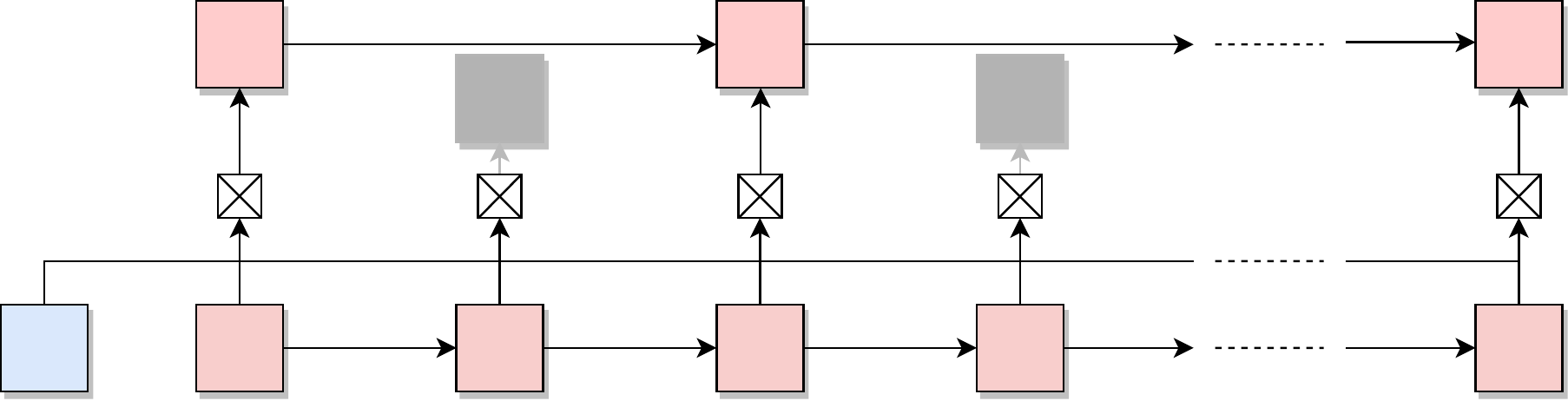}
    \vspace{-0.00cm}
    \caption{A visualization of FHE. Lower-Layer LSTM processes each step. For each token, the boundary gate decides (based on the current lower-layer LSTM state and question embedding) if information should be stored in the upper-level representation. Higher-Layer LSTM states update only when the corresponding gate is open.}
    \label{fig:model}
\vspace{-0.00cm}
\end{figure}

Our model consists of: the \emph{focused hierarchical encoder} (FHE), the context encoder and the decoder. Compared to a regular RNN with attention, we replace the encoder with a context-aware focused RNN encoder.

The \emph{focused hierarchical encoder} is modeled by a  two-layer LSTM. The lower layer operates at the input token level, while the upper layer focuses on tokens relevant to the context. We train a conditional boundary gate to decide, depending on the context or question, whether it is useful to update the upper-level LSTM with a summary of the current set of tokens or not.

\paragraph{Lower-level Layer} As shown in Figure~\ref{fig:model}, FHE has two layers.  Let $P = (\mathbf{x}_1, \ldots, \mathbf{x}_n)$ be the sequence of input tokens, $h_t$ be the LSTM hidden state and $c_t$ be the LSTM cell state at time $t$. To make our model as generic as possible, the lower-level layer may be also augmented with other available information. The question for large QA tasks are non-trivial and hence we augment the  lower-layer inputs with the question encoding at each step.
\begin{equation}
    \mathbf{h}_t^l, \mathbf{c}_t^l = \textmd{LSTM}(\mathbf{x}_t, \mathbf{h}^l_{t-1}, \mathbf{c}^l_{t-1})
\end{equation}

\paragraph{Conditional Boundary Gate} 
For each token in the passage, the boundary gate decides if information at the current time step should be stored in the upper-level representation.
We hypothesize that the question is essential in deciding how to represent the passage. To capture this dependency, the boundary gate computation is conditioned on the question embedding $\mathbf{q}$. The question embedding can vary in complexity depending on the difficulty of the task. In the simplest setting, the question embedding is simply a retrieved vector.

The output of the boundary gate is a scalar $b_t \in (0,1)$ that is taken to be the parameter of a Bernoulli distribution $\tilde b_t \sim \textmd{Bernoulli}(b_t)$ that regulates the gate's opening at time step $t$. In the simplest case, the boundary gate forward pass is formulated as 
\begin{equation}
    \label{eq:boundaries}
    b_t = \sigma(\mathbf{w}_b^\top\,\textmd{LReLU} (\mathbf{W}_b \mathbf{z}_t + \mathbf{b}_{b})),
\end{equation}
where $\mathbf{W}_b$, $\mathbf{b}_b$ and $\mathbf{w}_{b}$ are trainable weights, $\textmd{LReLU}(\cdot)$ is a leaky ReLU activation,  and $\mathbf{z}_t$ is the input that varies and depends on the task. In our experiments 
we used the following input
\begin{equation}
\mathbf{z}_t = [\mathbf{q} \odot \mathbf{h}^l_t,  \mathbf{h}^l_t, \mathbf{q}],
\end{equation}
where $\odot$ is the element-wise product. Hence, we essentially use three groups of features: question representation multiplied with lower-layer hidden states, question, and lower-layer representations. These three groups are concatenated together and passed through an MLP to yield boundary gate decisions (i.e., open/close).

In a more complex task that has stronger dependency on the upper-level hidden states (see the following paragraph), these can also be used to augment the boundary gate input $\mathbf{z}_t$.

\paragraph{Upper-level Layer}
The upper-layer LSTM states (denoted $\mathbf{h}_t^u$) update only when the corresponding lower-layer boundary gate is open.
\begin{eqnarray}
    \mathbf{\tilde h}_t^u,\mathbf{\tilde c}_t^u &=& 
    \textmd{LSTM}(\mathbf{h}^l_t, \mathbf{h}^u_{t-1}, \mathbf{c}^u_{t-1}) \\
    \tilde{b}_t &\sim& \textmd{Bernoulli}(b_t) \\
     \mathbf{c}^u_t &=& \tilde b_t \mathbf{\tilde c}_t^u + (1 - \tilde b_t) \mathbf{c}_{t-1}^u \\
    \mathbf{h}^u_t &=& \tilde b_t \mathbf{\tilde h}_t^u + (1 - \tilde b_t) \mathbf{h}_{t-1}^u
\end{eqnarray}

\paragraph{Final Output}
The final output of FHE is a sequence of lower-level states $H^l = \{\mathbf{h}^l_1, \ldots, \mathbf{h}^l_n\}$ and a sequence of upper-level states $H^u = \{\mathbf{h}^u_1, \ldots, \mathbf{h}^u_n\}$, where only $k$ of them are unique and $k = \sum_t \tilde{b}_t$, the number of times the boundary gates open over the length of the document. The upper-level states $H^u$ are typically the only ones being processed by the downstream modules. Hence, all downstream operations are performed faster than if they had to process $H^l$ (the effective size of $H^u$ is smaller than the size of $H^l$).

\subsection{Training}\label{subsec:training}
We train our model to maximize the log-likelihood of the answer ($A$) given the context ($Q$) and the passage ($P$), using the output probabilities given by our answer decoder:
\begin{equation}
\mathcal{R} = \log p(A \mid Q, P).
\end{equation}

\paragraph{Policy Gradient}
The discrete decisions involved in sampling the boundary variable $\tilde{b}_t$ make it impossible to use standard gradient back-propagation to learn the parameters of the boundary gate. We instead apply REINFORCE-style estimators \citep{williams1992}. Denote by $\pi_b$ the model policy over the binary vector of decisions $\mathbf{b} = \{\tilde b_1, \ldots, \tilde b_n\}$. We need to take the derivative:
\begin{equation}
\sum_{\mathbf{b}} \nabla \pi_b(\mathbf{b}) \mathcal{R}_\mathbf{b} = E_{\mathbf{b}\sim \pi_b}[\nabla \log \pi_b(\mathbf{b}) \mathcal{R}_\mathbf{b}],
\end{equation}

where the reward $\mathcal{R}_\mathbf{b}$ can be formulated differently depending on the task. In our synthetic experiments (Section \ref{sec:toy}), we let $\mathcal{R}_\mathbf{b} = \frac{\partial h^u}{\partial \tilde b_t}$. For large scale natural language QA tasks (Section \ref{sec:large_sacelQA}), we use $\mathcal{R}_\mathbf{b} = \log p(A \mid Q, P, \mathbf{b})$. The aforementioned gradient can be approximated by sampling from the policy $\pi_{b}$ and computing the corresponding terms.

\paragraph{Rewards}
We use the final reward $\mathcal{R}_\mathbf{b}$ for each decision in the sequence. Following previous work~\citep{williams1992,andrychowicz2016learning}, we add an exploration term $\alpha H(\pi_b)$ that prevents the policy from collapsing too soon during training. The $\alpha$ is a hyperparameter to be set.

\paragraph{Sparsity Constraints}
We add a constraint on the sparsity of the upper-level representations. We want the model to avoid grouping each token on its own and storing information at each step on the upper level (i.e., always opening the boundary gates). As a remedy we add a small penalty $G(\mathbf{b})$ the model needs to pay for storing information at the upper level. In practice, we found the following formulation to work the best:
\begin{equation}
\beta G(\mathbf{b}) = \beta ReLU\left( \big(\sum_{t=1}^{T}{b}_t \big) - \gamma T\right)
\end{equation}
where $\beta > 0$ and $\gamma \in [0,1)$ are hyper-parameters and $T$ is the input sequence length.
Hence, $\beta$ is the strength of penalty and $\gamma$ is the proportion of the time the gates could open without being penalized. Intuitively, we let a certain number of gates until a open threshold $\gamma T$  without any penalty. Each open gate above the threshold is penalized.  This is the same as constraining the policy to act within a certain region. One can skip $\gamma$ (by setting $\gamma = 0$) and then the penalty is just $\beta b_t$ applied at each time step:
\begin{equation}
\beta G(\mathbf{b}) = \beta ReLU \left(\big( \sum_{t=1}^{T}{b}_t\big) - \gamma T \right) \overset{\mathrm{\gamma = 0}}{=} \sum_{t=1}^{T} \beta b_t.
\end{equation}
Note that hyper-parameters $\beta$ and $\gamma$ directly affect the sparsity of upper-level representations that can be formally defined as the average value of $\tilde{b_t}$ and will be called \emph{gate openness}.

\section{Related Work}
As we have discussed above, our \emph{focused hierarchical encoder} is modeled by a hierarchical RNN controlled by gates conditioned on an input context or question. 
The idea of using hierarchical RNNs to model data in which long term dependencies must be captured was first explored in \citet{el1995hierarchical}.  

More recently, \citet{koutnik2014clockwork} propose a stacked RNN with a different updating rate for each layer, fixed a priori. \citet{graves2016adaptive} propose a RNN that learns the number of timesteps to ponder on an input before moving onto the next input. \citet{srivastava2015highway} utilizes skip-connections between layers in a feedforward network for training a deep network. \citet{yao2015depth} uses a soft differential depth gate to connect the lower and the upper layers and \citet{sordoni2015hierarchical} explore a multi-scale architecture where the hierarchy is fixed. Both of these uses a soft-differential gate compared to what can be seen as a hard gate in the \citet{chung2016hierarchical}. The Skip-RNN \cite{campos2017skip} learns an updating rate by predicting how many steps to skip in the future. Our document encoder bears similarities to the Hierarchical Multi-Scale LSTMs  (HMSTMs) of \citet{chung2016hierarchical}. The HMLSTM extends a 3-layered LSTM to have multiple gates at each time step, which decide which of the LSTM layers should be updated, and has been applied to unconditional character-level language modelling. In contrast we learn a context-conditional sequence segmentation that only encodes relevant information to the context; this information is fed to an attention mechanism to help with identifying the most relevant information.  

The upper states of our network can be considered as a \emph{memory} focusing on relevant information that can be attended to at each step of the answer generation process. In particular, we use a soft-attention mechanism \cite{bahdanau2014}, which has become ubiquitous in conditional language generation systems. \citet{yang2016hierarchical} and \citet{kumar2016ask} use a layered, hierarchical attention in that they attend to both word and sentence level representations. We use similar ideas but we learn how to attend to information within the sequence structure rather than relying on a fixed strategy. Another form of structured encoding encoding mechanism would be the \citet{miller2016key}, where the attention is separate into pairs of key and value. The key corresponds to the attention distribution and the value is used to encode the context.

\section{Synthetic Experiments}\label{sec:toy}
We first study two synthetic tasks that allow us to analyze our proposed gating, attention mechanism and its generalization ability, and then in Section \ref{sec:large_sacelQA} we study the more complex tasks of natural language question and answering.

The synthetic tasks are the \emph{picking task} and the \emph{Pixel-by-Pixel MNIST QA task}. For the \emph{picking task}, we analyze the gating mechanism and show how the model utilizes the question (context) to dynamically group the passage tokens and how the attention mechanism utilizes this information. We also test the generalization ability our model following the setup in \cite{graves2014neural}. For the \emph{Pixel-by-Pixel MNIST QA task}, we show better accuracy with our FHE module over the baseline. The  tasks are chosen due to the natural of the tasks. The gating mechanism for the \emph{picking task} depends solely on the question, whereas the gating mechanism for the \emph{Pixel-by-Pixel MNIST QA task} is independent of the question, but solely dependent on the data.

We compare the performance of our \emph{focused hierarchical encoder} module to two baseline architectures: a 1-layer LSTM (LSTM1) and a 2-layer LSTM (LSTM2)\footnote{Note that baseline models are equivalent to FHE with the boundary gate fully open for LSTM2 ($b_t = 1$ for each $t$) or always closed for LSTM1 ($b_t = 0$ for each $t$).}. 

For the \emph{picking task}, FHE utilizes less memory compare to LSTM2, as the baseline LSTM2 model needs to store and attend over all states, whereas FHE only needs to attend to unique elements of $H^u$. For example, when gate openness is below $10\%$, the attention module for FHE only attends to than $10\%$ of memory compared to a LSTM2 baseline model.

\begin{table}[ht!]
\vskip -0.1in
\caption{Sample points for \emph{picking task} (sequence length \mbox{$n=30$}). The first $k$ digits are underlined and the target mode is bolded.}
\label{table:pickingSample}
\vskip 0.1in
\begin{center}
\begin{small}
\begin{sc}
\begin{tabular}{l c | c} 
\toprule
\multicolumn{2}{c|}{\textbf{Input}} & \textbf{Target}\\
\textbf{Sequence} & \textbf{k} & \textbf{Mode}\\
\midrule
\textit{random examples} &&\\
\underline{8\textbf{0}56\textbf{0}2\textbf{0}17\textbf{0}}82838371701316304473 & 10 & \textbf{0}\\
\underline{6\textbf{3}87\textbf{33}290890\textbf{3}966902559\textbf{3}}7986485 & 23 & \textbf{3}\\
\underline{\textbf{1}6455\textbf{1}9375793738968\textbf{1}398\textbf{11}2}5982 & 26 & \textbf{1}\\
\midrule
\textit{malicious examples} &&\\
\underline{\textbf{666}333}666288882888819999999990 & 6 & \textbf{6}\\
\underline{\textbf{666}333\textbf{666}2}88882888819999999990 & 10 & \textbf{6}\\
\underline{6663336662\textbf{8888}2\textbf{8888}1}9999999990 & 20 & \textbf{8}\\
\underline{66633366628888288881\textbf{999999999}}0 & 30 & \textbf{9}\\
\bottomrule
\end{tabular}
\end{sc}
\end{small}
\end{center}
\vskip -0.25in
\end{table}

\paragraph{Hyper-parameters} All models (FHE, LSTM1 and LSTM2) for a certain task has the same number of hidden units (256 for \emph{picking task} and 128 for \emph{Pixel-by-Pixel MNIST QA task}).
In FHE module we used $\alpha=0$ hence, we did not use exploration term mentioned in Section \ref{subsec:training}. Instead, we used simpler idea that is sufficient in the  synthetic experiments conducted -- we add a small value to $b_t$ (0.01 for \emph{picking task} and 0.1 for \emph{Pixel-by-Pixel MNIST QA task}) to encourage exploration. The values of $\beta$ and $\gamma$ depend on the task and are provided later. Learning rates used for all models are 0.0001 with the Adam optimizer \cite{kingmaandba2014}.

\subsection{Picking task}

Given a sequence of randomly generated digits of length $n$, the goal of the \emph{picking task} is to determine the most frequent digit within the first $k$ digits\footnote{If there is more than one  mode, the largest value digit should be picked.}, where $k \leq n$. Hence, the value of $k$ is understood as the question. We study three tasks with input sequences of $n \in \{100, 200, 400\}$ digits respectively. Sample points for the task are presented in Table~\ref{table:pickingSample}.

The input digits $\mathbf{x}_i$ are one-hot encoded vectors (size 10) and the question embedding $\mathbf{q}$ is a vector retrieved from the lookup table (that is learnt during training) with $n$ entries. To obtain the final sequence representation, soft attention (as in \citet{bahdanau2014}) is applied on the upper-level states $H^u$ (for FHE and LSTM2) or the lower-level states $H^l$ (for LSTM1). Finally, the representation is concatenated with the question embedding and fed to one layer feed-forward neural network to produce the final prediction (i.e.\, probabilities for all 10 classes).

As introduced in Section~\ref{subsec:training}, there are two hyper-parameters ($\beta$ and $\gamma$) that affect the sparsity of higher-level representations in FHE. We explore two approaches for determining their values.

\begin{figure}[hpt!]
\begin{minipage}{0.48\textwidth}
\centering
\includegraphics[width=0.9\linewidth]{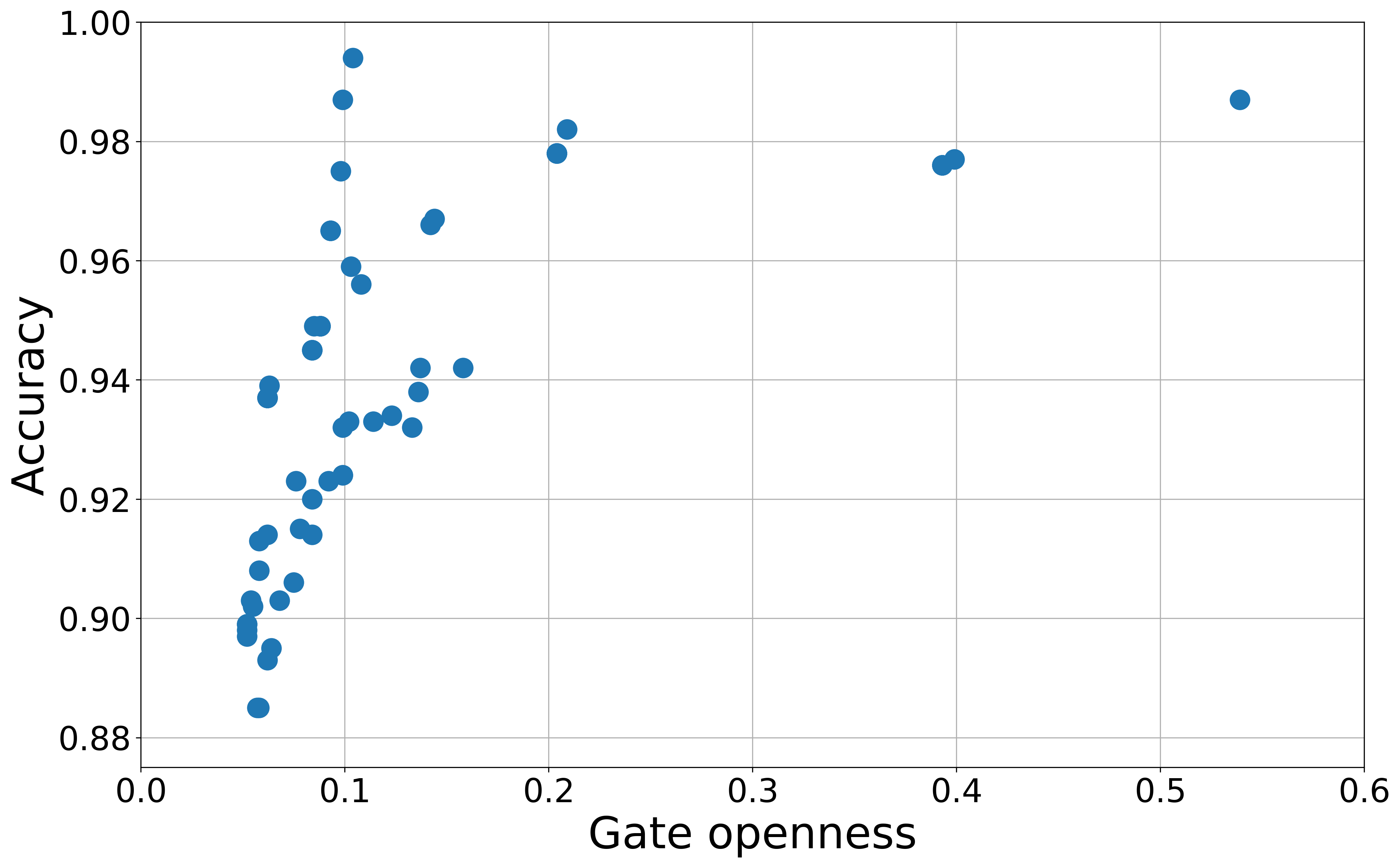}
\vspace{-0.00cm}
\caption{A relationship between accuracy and gate openness for \emph{picking task} and sequence length $n =100$. The best performance is achieved for gate openness around $10\%$.}
\label{accVSopen100}
\end{minipage}

\bigskip

\begin{minipage}{0.48\textwidth}
\centering
\includegraphics[width=0.81\linewidth]{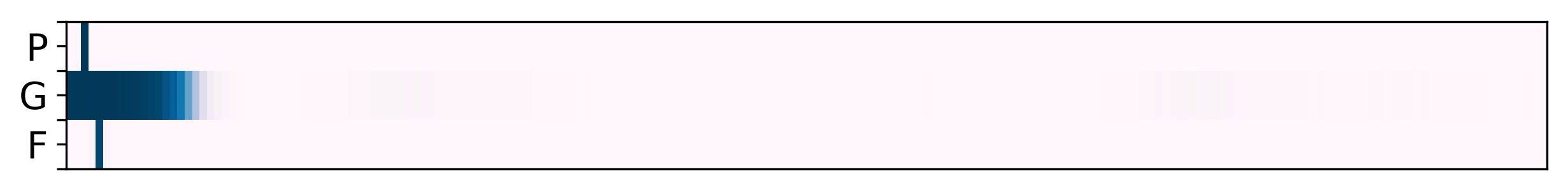} 
\includegraphics[width=0.81\linewidth]{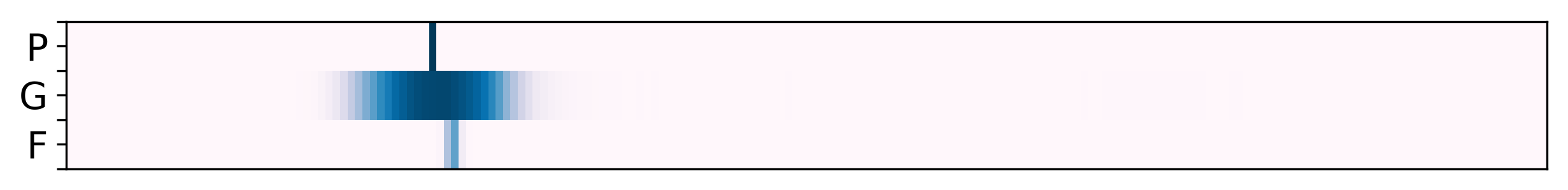}
\includegraphics[width=0.81\linewidth]{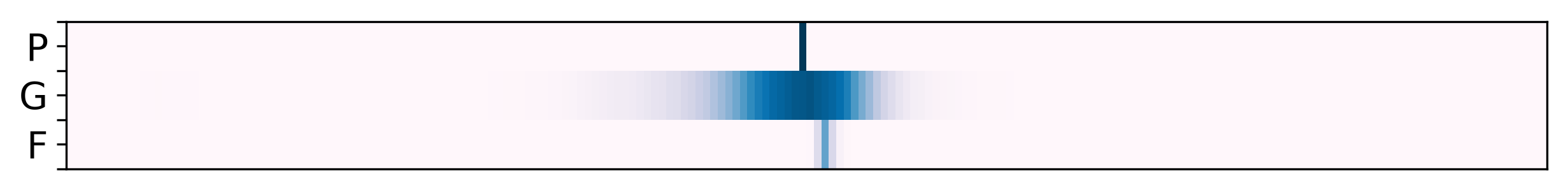} 
\includegraphics[width=0.81\linewidth]{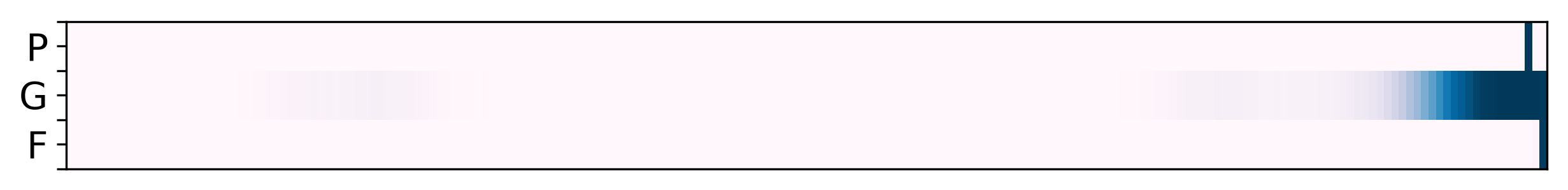} 
\includegraphics[width=0.81\linewidth]{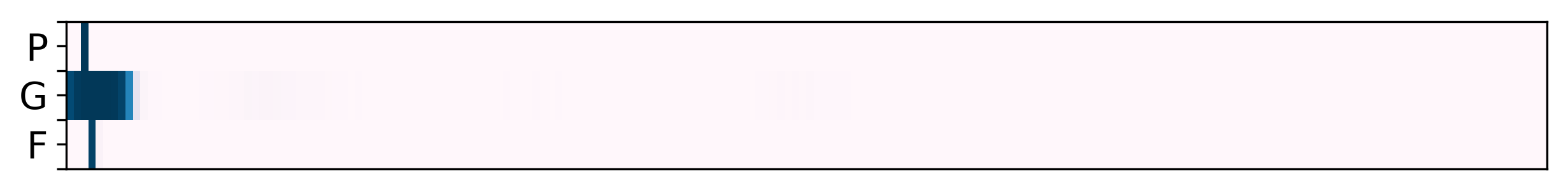} 
\includegraphics[width=0.81\linewidth]{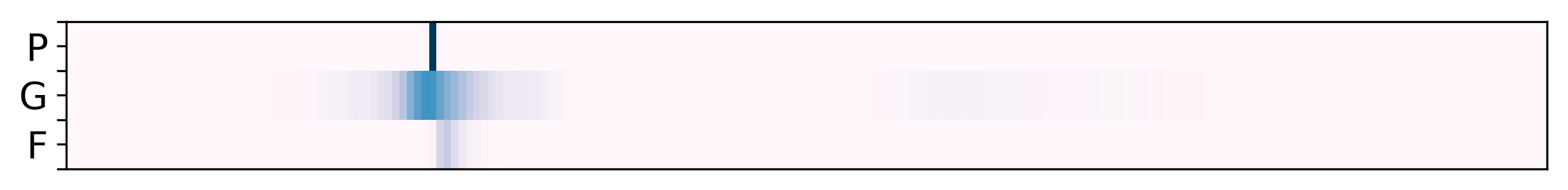}
\includegraphics[width=0.81\linewidth]{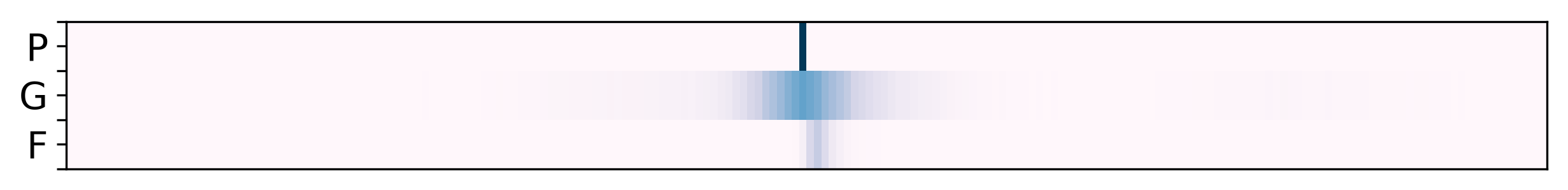} 
\includegraphics[width=0.81\linewidth]{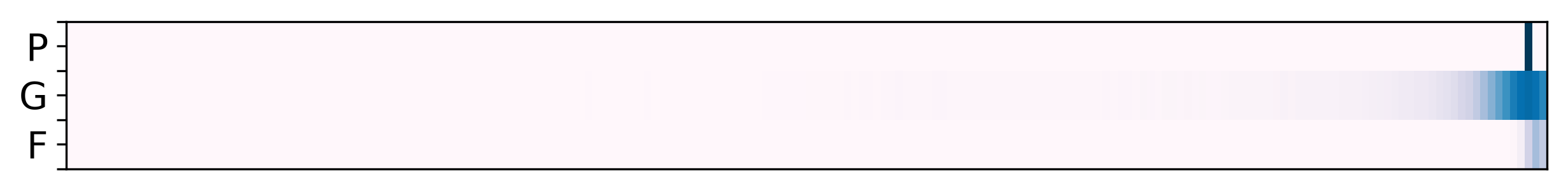} 
\vspace{-0.00cm}
\caption{Gate openness (G) conditioned on the position asked (P). Focus (F) is the average of final attention weight set for a given step. Hence, focus sums to one and it is always lower than gate openness (because our model attends only over unique states). Result showed for sequence length $n = 200$. The first four plots illustrate FHE model having $99.4\%$ accuracy and $10\%$ gate openness, while the last four are for FHE model having $97\%$ accuracy but $5\%$ gate openness.}
\label{fig:openness}
\end{minipage}

\bigskip

\begin{minipage}{0.48\textwidth}
\centering
\includegraphics[width=0.8\linewidth]{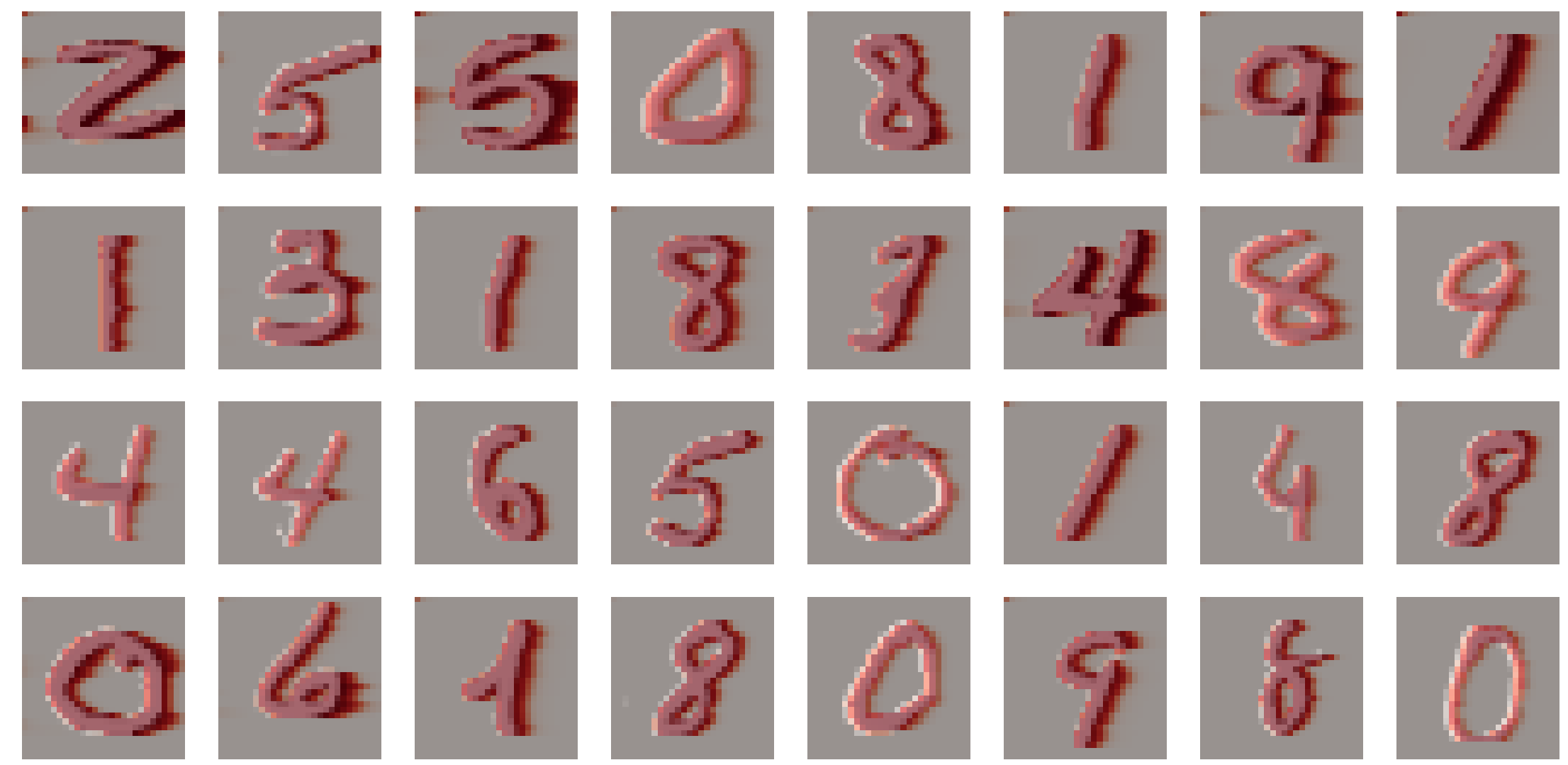}
\caption{A visualization of the gating mechanism learned using the Pixel-by-Pixel MNIST dataset. Red pixels indicate a gate opening and are overlayed on top of the digit which is white on a gray background. The digits are vectorized row-wise which explains why white pixels appear left of the red pixels.}
\label{fig:qca}
\vspace{-0.00cm}
\end{minipage}
\end{figure}

One approach is to fix these hyper-parameters to a small value (for example  $\beta = 0.1$ and $\gamma = 0.25$) in the beginning of training, such that the gates can almost freely open. Once the desired accuracy has been reached, we enforce constraints on our hyper-parameters. This provides a level of control over the accuracy-sparsity trade-off -- we used this approach  with the requirement of achieving a desired accuracy $a$. We tested FHE models with \mbox{$a \in \{80\%, 90\%, 95\%, 98\%\}$} and call them FHE80, FHE90, etc. The relationship between accuracy and gate openness is visualized in Figure \ref{accVSopen100}. 

Another approach is to set $\beta$ and $\gamma$ to a fixed value from the start, so the gate openness of the model is more restricted right from the start. We find that the model performs better with fixed the hyper-parameters (the results for $\beta = 1$ and $\gamma = 10\%$ are presented in Table \ref{table:picking} as FHE-fixed).

The results achieved for each model and sequence length are presented in Table \ref{table:picking} and Table \ref{table:pickingCHAa}. For each setup at least two runs were performed and the difference in result between the pair were typically neglectable ($<0.5\%$).

\begin{table}[t]
\vskip -0.1in
\caption{Accuracy ($\%$) for \emph{picking task} for LSTM1, LSTM2 and FHE-fixed. Our model and LSTM2 are on par with performing while LSTM1 is behind for longer input sequences.}
\label{table:picking}
\vskip 0.1in
\begin{center}
\begin{small}
\begin{sc}
\begin{tabular}{c || c c | c} 
\toprule
\textbf{Length} & \textbf{LSTM1} & \textbf{LSTM2} & \textbf{FHE-fixed}\\
\midrule
100 & 99.4 & \textbf{99.7} & 99.5\\
200 & 97.0 & 99.2 & \textbf{99.4}\\
400 & 92.9 & \textbf{97.5} & 96.9\\
\bottomrule
\end{tabular}
\end{sc}
\end{small}
\end{center}

\caption{Accuracy ($\%$) for \emph{picking task} for the models providing a level of control over the accuracy-sparsity trade-off at a cost of slightly lower performance.}
\label{table:pickingCHAa}
\vskip 0.1in
\begin{center}
\begin{small}
\begin{sc}
\begin{tabular}{c || c c c c}
\toprule
\textbf{Length} & \textbf{FHE80} & \textbf{FHE90} & \textbf{FHE95} & \textbf{FHE98}\\
\midrule
100 & 93.4 & 94.2 & 96.6 & \textbf{98.7}\\
200 & 92.3 & 92.4 & \textbf{93.6} & \textbf{93.6}\\
400 & 87.2 & 90.5 & 90.0 & \textbf{91.0}\\
\bottomrule
\end{tabular}
\end{sc}
\end{small}
\end{center}

\caption{Test accuracy ($\%$) for longer sequence length for \emph{picking task} on model trained on sequence length \mbox{$n = 200$}.}
\label{table:pickingGeneralization}
\vskip 0.1in
\begin{center}
\begin{small}
\begin{sc}
\begin{tabular}{c || c c | c} 
\toprule
\textbf{Length} & \textbf{LSTM1} & \textbf{LSTM2} & \textbf{FHE-fixed}\\
\midrule
200 & 97.1 & 99.2 & \textbf{99.4}\\
400 & 55.9 & 61.4 & \textbf{97.6}\\
800 & 39.6 & 39.7 & \textbf{95.6}\\
1600 &  29.5 & 28.6  & \textbf{93.3}  \\
10000 &  18.5 & 14.8 & \textbf{66.8} \\
\bottomrule
\end{tabular}
\end{sc}
\end{small}
\end{center}
\vskip -0.25in
\end{table}

The \emph{picking task} is useful to validate our gating mechanism. Once trained we can inspect the positions of the opened gates. Figure~\ref{fig:openness} shows that our model learns to open gates around $k$'th step only and attend a single gate right after the $k$'th step. The lower-level LSTM is used to count the occurrences of the various digits. That information is then passed to the upper-level LSTM at a single gate. The attention mechanism then uses the information around the same time step to provide the mode (i.e.\, solve the task).

We tested the generalization ability of FHE. The models trained on short sequences (\mbox{$n=200$}) were evaluated on longer sequences and \mbox{$k\leq200$}. The results are in Table~\ref{table:pickingGeneralization}. The models can not be evaluated for $k$ larger than maximum sequence length used during training because the question embeddings are parts of the models. FHE generalizes better to longer sequences by a wide margin. We believe this is due to the boundary gates being open for only the first k-steps, and the attention mechanism not attending over possibly misleading states.

\subsection{Pixel-by-Pixel MNIST QA task}

\begin{table}[t!]
\vskip -0.1in
\caption{Accuracy (\%) for validation set of \emph{Pixel-by-Pixel MNIST QA task}. Our model slightly outperform both LSTM1 and LSTM2.}
\label{table:mnist}
\vskip 0.1in
\begin{center}
\begin{small}
\begin{sc}
\begin{tabular}{c c | c}
 \toprule
 \textbf{LSTM1} &  \textbf{LSTM2} & \textbf{FHE-fixed}\\
 \midrule
 97.3 & 98.4 &  \textbf{99.1}\\
 \bottomrule
\end{tabular}
\end{sc}
\end{small}
\end{center}
\vskip -0.25in
\end{table}

We adapt the Pixel-by-Pixel MNIST classification task \cite{lecun1998gradient,DBLP:journals/corr/LeJH15} to the question and answering setting. The passage encoder reads in MNIST digits one pixel at a time. The question asked is whether the image is a specific digit and the answer is either \emph{True} or \emph{False}. The data is balanced such that approximately half of the answers are \emph{True} and the other half are \emph{False}.

The LSTM2 reached an accuracy of $98.4\%$ on the validation set, and FHE\footnote{We used $\beta=0.0001$ and $\gamma=50\%$.} outperformed the baseline by having an accuracy of $99.1\%$. Figure \ref{fig:qca} shows a visualization of the gates for the passage encoder learned by the model. The model learns to open the boundary gate almost always around the digit. We also found that for this particular task, the gates do not depend on the question. We hypothesize that this is because it is much easier for the passage encoder to learn to open the gates when there is a white pixel. In any case, these experiments illustrate how our proposed mechanism  modulates gates based on input questions and features in the data.

\section{Large Scale Natural Language QA Tasks}\label{sec:large_sacelQA}
Next, we explore the more complex task of natural language question answering. We study our approach using the MS MARCO and SearchQA datasets and tasks. These tasks are well-suited for our model since they both involve searching over a long input passage for answers to a question. Our results are that for the MS MARCO task, we achieved scores higher than the baseline models. Our model on SearchQA significantly outperforms very recent work \cite{buck2018ask}. We also run ablation studies on the model for MS MARCO task to show the importance of each component in the model.

To obtain competitive results on these difficult question-answering tasks we embed FHE with a modified version of both the question encoder and the answer decoder. All changes with respect to what was presented earlier are detailed in the following sections.

\subsection{Question Encoder}
Following recent work~\cite{cui2016attention,chen2017reading}, we use a bidirectional LSTM that first reads the question and then performs self-attention to construct a vector representation of it. At the model-level, the question-encoder module outputs the vector $\mathbf{q}$, which is then used as conditioning information in a FHE.

\subsection{Decoder} \label{sec:answer_decoder}
The answer decoder  follows the standard decoding procedure in RNN with attention \cite{bahdanau2014,gulcehre2016pointing}. The only difference is that the decoder looks over the upper-level hidden states $h^u_t$ learned using a FHE conditioned on the question. 
The upper-level states $H^u$ provide an abstracted, dynamic representation of the passage. Because they receive lower-layer input only when the boundary gate is open, the resulting hidden states can be viewed as a sectional summary of the tokens between these ``open'' time-steps. The upper layer thus summarizes a passage in a smaller number of states. This can be beneficial because it enables the encoder LSTM to maintain information over a longer time-horizon, reduces the number of hidden states, and makes learning the subsequent attention softmax layer easier.

\paragraph{Pointer Softmax} In order to predict the next answer word and to avoid large-vocabulary issues, we use the pointer softmax~\cite{gulcehre2016pointing}. This method decomposes as two softmaxes: one places a distribution over a shortlist of words and the other places a distribution over words in the document. The softmax parameters are $W_o \in R^{|V| \times D_h}$ and $b_o \in R|V| $, where $|V|$ is the size of the shortlist vocabulary\footnote{We use a short-list of 100 or 10,000 most frequent words for SearchQA or MS MARCO tasks, respectively.}.  A switching network enables the model to learn the mixture proportions over the two distributions. Switching variable $z_j$ determines how to interpolate between the indices in the document and the shortlist words. It is computed via an MLP. Let $o_j$ be the distribution of word in the shortlist, and $\alpha_j$ be the distribution over words in the document index. Then the  pointer softmax $P_j \in R^{|V|+D}$ is $P_j = [z_j o_j; (1-z_j)\alpha_j]$. 

\begin{table*}[ht!]
\vskip -0.1in
\caption{SearchQA results measured in F1 and Exact Match (EM) for validation and test set. Our model and AMANDA ~\cite{kundu2018question} are on par with performing while the other models are behind.}
\label{tab:searchqa_results}
\vskip 0.1in
\begin{center}
\begin{small}
\begin{sc}
\begin{tabular}{l||cc|cc}
\toprule
\multirow{2}{*}{\textbf{Models}} & \multicolumn{2}{c|}{\textbf{Validation}} & \multicolumn{2}{c}{\textbf{Test}} \\
& \textbf{F1} & \textbf{EM} & \textbf{F1} & \textbf{EM} \\
\midrule
TF-IDF Max~\cite{dunn2017searchqa}  & - & 13.0 & - & 12.7\\
ASR ~\cite{dunn2017searchqa} & 24.1 & 43.9 & 22.8 & 41.3\\
AQA ~\cite{buck2018ask}  & 47.7 & 40.5 & 45.6 & 38.7\\
Human  ~\cite{dunn2017searchqa}  &-& - & 43.9 & -\\
\midrule
LSTM1 + pointer softmax & 52.8 & 41.9 & 48.7 & 39.7\\
LSTM2 + pointer softmax & 55.3 & 44.7 & 51.9 & 41.7\\
\midrule
Our Model & 56.7 & \textbf{49.6} & 53.4 & \textbf{46.8}\\
\midrule
\textbf{Concurrent work} & & & & \\
AMANDA ~\cite{kundu2018question} & \textbf{57.7} & 48.6 & \textbf{56.6} & \textbf{46.8}\\
\bottomrule
\end{tabular}
\end{sc}
\end{small}
\end{center}
\vskip -0.1in

\caption{MS MARCO results using BLEU-1 and Rouge-L evaluation. Our model clearly outperforms both standard memory networks and sequence-to-sequence models. In addition in both Bleu-1 and Rouge-L, we outperform  strong baselines. Available results for the first three methods are taken from their respective papers (hence the not available ones).
Ablation study results show that our model benefits the most from elementwise product between questions and context. The pointer softmax also gives a significant gain for performance.}
\label{tab:marco_results}
\vskip 0.1in
\begin{center}
\begin{small}
\begin{sc}
\begin{tabular}{l||cc|cc}
\toprule
\multirow{2}{*}{\textbf{Generative Models}} & \multicolumn{2}{c|}{\textbf{Validation}} & \multicolumn{2}{c}{\textbf{Test}} \\
& \textbf{Bleu-1} & \textbf{Rouge-L} & \textbf{Bleu-1} & \textbf{Rouge-L} \\
\midrule
Seq-to-Seq~\cite{nguyen2016ms}  & -    &  8.9  & -   & -  \\
Memory Network~\cite{nguyen2016ms} &  - &  11.9   & -   & - \\
Attention Model~\cite{higgins2017lstm}  & 9.3 & 12.8  & -  & -  \\
\midrule
LSTM1 + pointer softmax & 24.8 & 26.5 & 28 & 28  \\
LSTM2 + pointer softmax & 24.3 & 23.3 & 27 & 28   \\
\midrule
Our Model & \textbf{27.3} & \textbf{26.7} & \textbf{30} & \textbf{30} \\
\midrule
\textbf{Ablation study} & & & & \\
Our Model -- dot-product  between question and context & 18.5 & 19.3 & - & - \\
Our Model -- pointer softmax & 20.5 & 18.7 & - & - \\
Our Model -- learned boundaries & 23.5 & 24 & - & - \\
\bottomrule
\end{tabular}
\end{sc}
\end{small}
\end{center}
\vskip -0.25in
\end{table*}

\paragraph{Hyper-parameters} All components of the model (FHE, question encoder, decoder) in all  natural language QA experiments uses 300 hidden units. FHE hyper-parameters were fixed ($\alpha=0.001$, $\beta=0.5$, $\gamma=50\%$). We use the Adam optimizer \cite{kingma2014} with a learning rate of 0.001.

\subsection{SearchQA Question and Answering Task}

Search QA \cite{dunn2017searchqa} is  large scale QA dataset in the form of Question-Context-Answer. The question-answer pairs are real Jeopardy! questions crawled from J!Archive. The contexts are text snippets retrieved by Google. It contains $140,461$ question-context-answer pairs. Each pair is coupled with a set of $49.6 \pm 2.10$ snippets, and each snippet is $37.3 \pm 11.7$ tokens long on average. Answers are on average $1.47 \pm 0.58$ tokens long.

We use the same metric as reported in \citet{buck2018ask}, which are  F1 scores for multi-word answers and Exact Match (EM) for single word answers.

\paragraph{QA Results} Our model outperformed the recently proposed AQA model \cite{buck2018ask} by 8 points in EM and more than 6 points in terms of F1 scores. See detailed results in Table \ref{tab:searchqa_results}.

\subsection{MS MARCO Question and Answering Task}

The Microsoft Machine Reading Comprehension Dataset (MS MARCO) \cite{nguyen2016ms} is one of the largest  publicly available QA datasets. Each example in the dataset consists of a query, several context passages retrieved by the Bing search engine (ten per query on average), and several human generated answers (synthesized from the given contexts). 

\begin{figure*}[ht!]
\centering
\small
\includegraphics[width=0.95\textwidth]{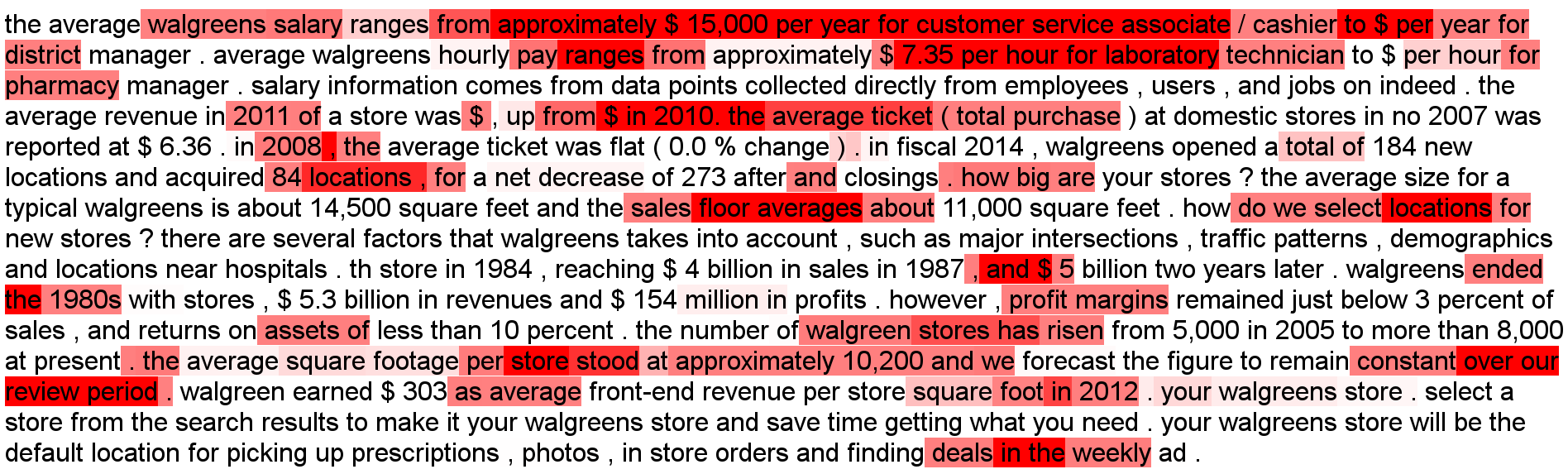}
\caption{We visualize the attention heat map of the passage encoder using an example from the MS MARCO validation set. Darker background color indicates higher attention (figure better seen on-screen). The question related to this passage is ``\emph{walgreens store sales average}'' and the ground truth answer is ``\emph{Approximately \$15,000 per year.}''. Our model learns to attend to the passage containing the answer (first line). In addition, most of the other high-intensity passages are related to the question.}
\label{fig:VW}
\end{figure*}

\paragraph{Span-based vs Generative}
Most of the recent question and answering models for MS MARCO are \emph{span-based}\footnote{For MS MARCO, span-based models are trained using ``gold-spans'', obtained by a preprocessing step which selects the passage in the document maximizing the Bleu-1 score with the answer.}  \cite{weissenborn2017making,tan2017s,shen2017reasonet,wang2016machine}. Span-based models are currently state of the art according to Bleu-1 and Rouge scores on the MS MARCO leaderboard, but are clearly limited as they cannot answer questions where the answer is not contained in the passage. In comparison, generative models, such as ours synthesize a novel answer for the given question. Generative models could learn a disentangled representation, and therefore generalize better.  Our approach takes the first step towards closing the gap between generative models and span-based models. 

We report the model performance using Bleu-1 and Rouge-L, which are the standard evaluation for MS MARCO task.

\paragraph{QA Results}
Table \ref{tab:marco_results} reports performance evaluated using the MS MARCO dataset. Specifically, we evaluate the quality of the generated answers for different models. Our model outperforms all competing methods in terms of test set Bleu-1 and Rouge-L. 

\paragraph{Ablation Studies}

Table \ref{tab:marco_results} shows also the results of learning our model without some of its key components.
The ablation studies are evaluated on the validation set only.

The largest gain came from the elementwise-product between question and context. This result is to be expected, since it is difficult for the model to encode the appropriate information
without direct knowledge of the question. 

The pointer softmax is another important module of the model. The MS MARCO dataset contains many rare words, with around $90\%$ of words appealing less than 20 times in the dataset. It is difficult for the model to generate words it has only seen a few times, and therefore the pointer-softmax provides a significant gain. 

Our experiments also show the importance of learned boundaries.  
This results is supportive of our hypothesis that learned boundaries help with better document encoding, and therefore generates better answers.

Overall, the different components in our model are all needed to achieve the final score.

\paragraph{Model Exploration}
Figure \ref{fig:VW} reports the results of our attention mechanism on an example from the MS MARCO dataset. Our attention focuses on the relevant passage (the one that contains the answer) as well as other salient phrases of the passage given the question.

\paragraph{Human Evaluation}
We performed a  human evaluation study to compare answers generated by our model to answers generated by the LSTM1 baseline model in Table \ref{tab:marco_results}.  We randomly selected 23 test-set questions and their corresponding answers. The order of the questions are randomized for each questionnaire. We collected a total of 690 responses (30 volunteers each given 23 examples) where volunteers were shown both answers side-by-side and were asked to pick their preferred answers.
$63\%$ of the time, volunteers preferred the answers generated from our model. Volunteers are students from our lab, and were not aware of which samples came from which model.

\section{Conclusion}
We introduced a focusing mechanism for encoder recurrent neural networks and evaluated our approach on the popular task of natural-language question answering. Our proposed model uses a discrete stochastic gating function that conditions on a vector representation of the question to control information flow from a word-level representation to a concept-level representation of the document. We trained the gates with policy gradient techniques. Using synthetic tasks we showed that the mechanism correctly learns when to open the gates given the context (question) and the input (passage). Further, experiments on MS MARCO and SearchQA -- recent large-scale QA datasets -- showed that our proposed model outperforms strong baselines and in the case of SearchQA outperforms prior work.
\section*{Acknowledgments}

The authors would like to thank Anirudh Goyal for help in shaping this idea. The authors would also like to Alex Lamb, Olexa Bilaniuk, Julian Serban and Sandeep  Subramanian for their helpful feedback and discussions. We also thank IBM, Google, and the Canada Research Chairs program for funding this research and
Compute Canada for providing access to computing resources. Nan Ke would like to thank Microsoft Research for helping to fund this project as a part of the research internship at Microsoft Research in Montreal. Konrad \.Zo\l{}na would like to acknowledge the support of Applica.ai project co-financed by the European Regional Development Fund (POIR.01.01.01-00-0144/17-00). Zhouhan would like to thank AdeptMind by helping to fund a part of this research through his AdeptMind scholarship. The authors would also like to express debt of
gratitude towards those who contributed to theano over the years (as it is no longer maintained),
making it such a great tool.

\bibliographystyle{iclr2018_conference}
\bibliography{bibliography}

\end{document}